\pdfoutput=1
\documentclass[11pt]{article}

\usepackage[]{emnlp2021}

\usepackage{times}
\usepackage{latexsym}

\usepackage[T1]{fontenc}
\usepackage[utf8]{inputenc}

\usepackage{microtype}

\usepackage{graphicx}
\usepackage{subcaption}
\usepackage{amsmath}
\usepackage{mathtools}
\usepackage{amsfonts}

\usepackage{times}
\usepackage{latexsym}
\usepackage{amsmath}
\usepackage{url}
\usepackage{mathdots}
\usepackage{cancel}
\usepackage{color}
\usepackage{array}
\usepackage{multirow}
\usepackage{amssymb}
\usepackage{gensymb}
\usepackage{tabularx}
\usepackage{booktabs}
\usepackage{svg}
\usepackage{tablefootnote}
\usepackage{cleveref}
\usepackage{tabularx}
\usepackage{xcolor}
\usepackage{makecell}
\usepackage{enumitem}
\usepackage{sidecap}

\usepackage{floatrow}
\newfloatcommand{capbtabbox}{table}[][\FBwidth]

\newcommand{\tsmethod}{\textsc{TRUCE}}

\newcommand{\retrieval}{\textsc{NearNbr}}
\newcommand{\seqff}{\textsc{FcEnc}}
\newcommand{\seqffmulti}{\textsc{FcEnc}}
\newcommand{\seqlstm}{\textsc{LstmEnc}}
\newcommand{\seqlstmmulti}{\textsc{LstmEnc}}
\newcommand{\seqfft}{\textsc{FftEnc}}
\newcommand{\seqconv}{\textsc{ConvEnc}}
\newcommand{\seqconvmulti}{\textsc{ConvEnc}}

\title{Truth-Conditional Captioning of Time Series Data}

\author{Harsh Jhamtani \\ 
School of Computer Science \\ 
Carnegie Mellon University \\
\texttt{jharsh@cs.cmu.edu} \\ \And
Taylor Berg-Kirkpatrick \\
Computer Science and Engineering\\
University of California San Diego\\
\texttt{tberg@ucsd.eng.edu} \\
}

\date{}

\begin{document}
\maketitle

\begin{abstract}
In this paper, we explore the task of automatically generating natural language descriptions of salient patterns in a time series, such as stock prices of a company over a week. A model for this task should be able to extract high-level patterns such as presence of a peak or a dip. While typical contemporary neural models with attention mechanisms can generate fluent output descriptions for this task, they often generate factually incorrect descriptions. We propose a computational model with a truth-conditional architecture which first runs small learned programs on the input time series, then identifies the programs/patterns which hold true for the given input, and finally conditions on \emph{only} the chosen valid program (rather than the input time series) to generate the output text description. A program in our model is constructed from modules, which are small neural networks that are designed to capture numerical patterns and temporal information. The modules are shared across multiple programs, enabling compositionality as well as efficient learning of module parameters. The modules, as well as the composition of the modules, are unobserved in data, and we learn them in an end-to-end fashion with the only training signal coming from the accompanying natural language text descriptions. We find that the proposed model is able to generate high-precision captions even though we consider a small and simple space of module types. 
\end{abstract}

\section{Introduction}

\begin{figure}[t]
    \includegraphics[width=0.95\textwidth]{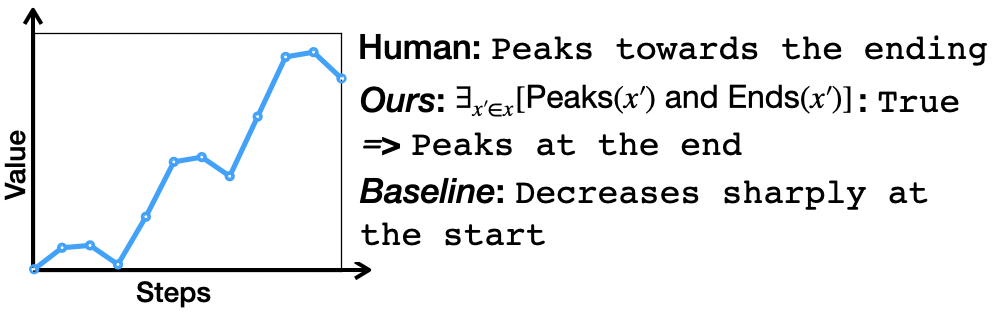}
    \vspace{-0.1\abovedisplayskip}
    \caption{\footnotesize We propose a neural truth-conditional model for high precision and diverse time series caption generation. 
    }
    \label{fig:pull}
    \vspace{-3mm}
\end{figure}

There has been a large interest in generating automatic text description \cite{mckeown1992text} of tabular data -- for example, prior work has sought to generate biographies from tables of biographical information \cite{lebret2016neural}, and generating descriptions from structured meaning representations \cite{gardent2017webnlg}. 
However, in many of these tasks, the main focus is on designing systems that are able to \emph{select entries} from tabular or equivalent data during generation by using neural attention mechanisms.
In many naturally occurring descriptions of tabular data, humans often refer to higher-level patterns, for example in the description of stock index pricing over the week in Fig. \ref{fig:pull}, 
the speaker refers to how the stock price peaks towards the ending. Some recent work has looked into setups that require non-trivial inference \cite{wiseman2017challenges,chen2020logical}. However, they typically don't involve inference about numerical patterns in time series data. 
Moreover, much recent prior work on identifying more complex patterns in data for captioning has relied on deep neural networks, often employing neural encoders and attention mechanisms. 
However, such approaches often fail to generate faithful responses and lack interpretability \cite{DBLP:journals/corr/abs-1910-08684,dhingra2019handling,parikh2020totto}. 

\begin{figure}[t]
\begin{center}
    \includegraphics[width=0.80\textwidth]{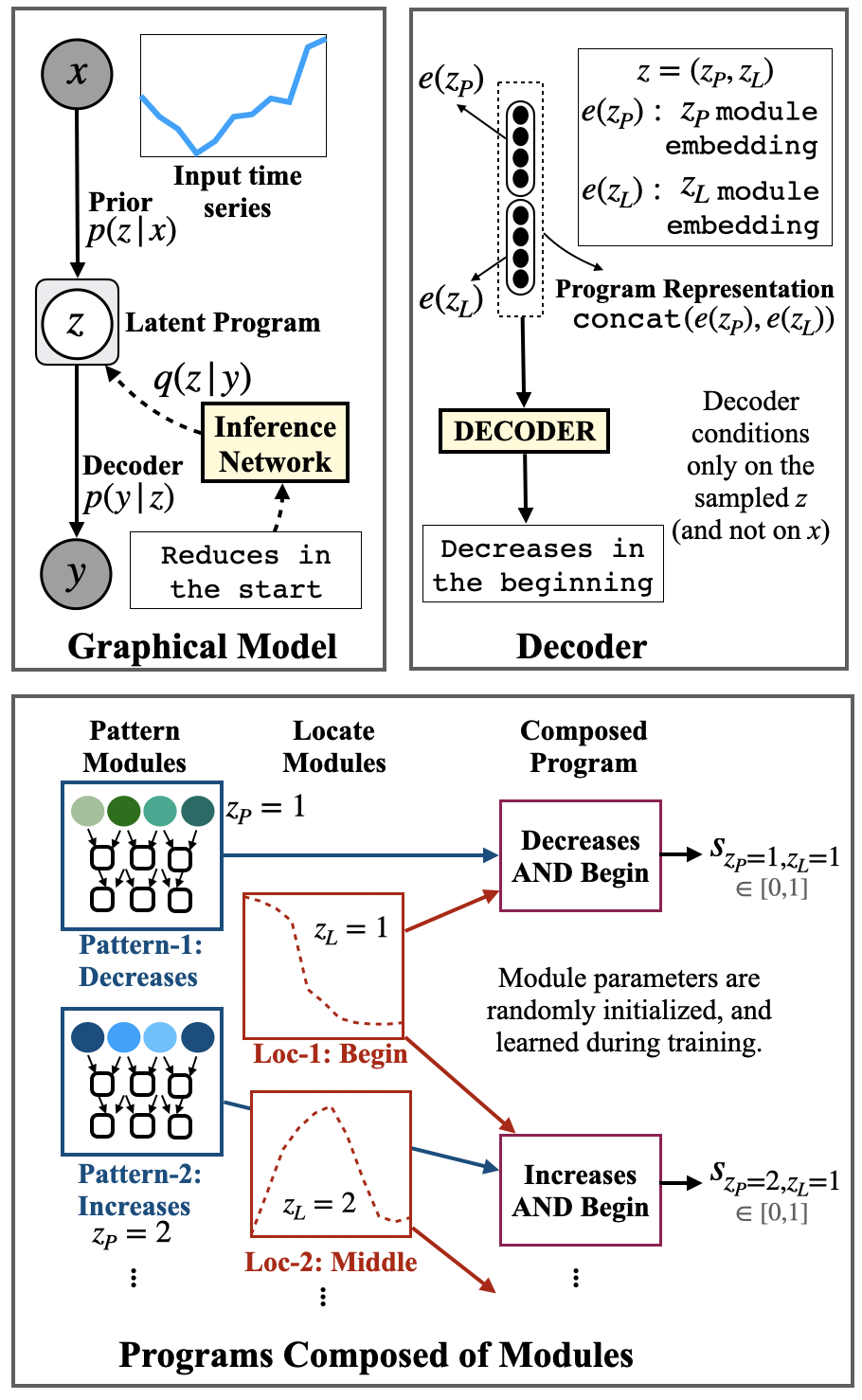}
    \vspace{-0.3\abovedisplayskip}
    \caption{\footnotesize Method Overview: We present a truth-conditional model for time series captioning, which first identifies patterns (composed of simpler modules) that hold true for a given data point. Decoder conditions only on a sampled program $z$ (and not on input $x$), generating high precision outputs. 
    } 
    \label{fig:ts_method_overview}
    \end{center}
    \vspace{-2mm}
\end{figure}

\begin{figure}[t]
\begin{center}
    \includegraphics[width=0.95\textwidth]{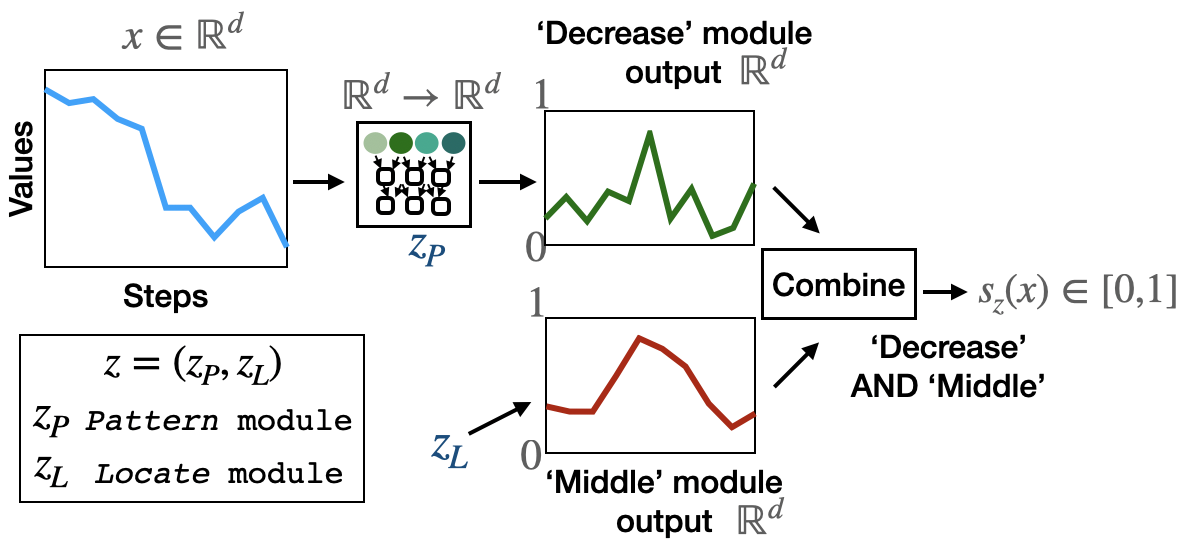}
    \vspace{-0.3\abovedisplayskip}
    \caption{\footnotesize 
    A program $z=(z_P,z_L)$ operates on an input time series $x$ to given final output score $s_z(x)$. The module instances are learned from scratch during training. 
    } 
    \label{fig:vizoutput}
    \end{center}
    \vspace{-2mm}
\end{figure}

We present a novel neural
truth-conditional model for time series captioning, which learns to identify patterns that hold true for the input time series (Figure \ref{fig:ts_method_overview}). 
We first sample a latent program from the space of learned neural operators. Each program produces a soft truth-value. Then, with probability proportional to each program's truth-value, a language decoder generates a caption. Thus, programs that yield low truth values, do not produce captions. Critically, the decoder takes \textit{an encoding of the program itself}, rather than the time series, in order to determine output text. Overall, this approach allows for both: (a) precision in generated output through explicit truth conditioning, and explicit program structure as a representation of time series trends, and (b) diversity in caption generation through the sampling process.

While some of the patterns in data are complex, they can be considered to have been constructed by composing simpler concepts such as slope (rate of change of value) or comparisons (between values at give points). As such, our programs are constructed by composing simpler operations/modules.
Such a modular design enables sharing of modules across multiple programs, leading to more data efficient learning of module parameters, and also providing better generalization to unseen compositions of modules. 
We consider a relatively simple space of three module types, using which our model is able to capture a significant fraction of the patterns present in data. The module types could be expanded in future to capture more complex patterns.
Our model treats the choice of composed computation graph of programs as a latent variable, learned using natural language descriptions as the only supervision. 
In this respect, our approach is related to neural module networks used in \citet{andreas2016learning,andreas2016neural}, which condition on a question to generate a program, which then operates on an image or other data to predict an answer. 
In our case, the constructed computation graph operates and identifies salient patterns in the source data directly, without being guided by an input question.

Our main contributions are as follows:
We propose a novel method for time series captioning which first induces useful patterns via composing simpler modules, identifies the programs which hold true, and finally generates text describing the selected program.
Towards this end, we collect and release two datasets consisting of time series data with accompanying English language description of salient patterns.
We observe that the proposed method is able to learn useful patterns, exhibits compositionality and interpretability, and generates outputs that are 
% much more precise 
much more faithful to the input 
compared to strong traditional neural baselines.
% \footnote{Data and code will be publicly released.}
\footnote{Data and code can be found at \url{https://github.com/harsh19/TRUCE}.}

\section{Truth-Conditional Natural Language Description} 

Our goal is to learn models for describing salient patterns in time series data. The main research challenge involved is to learn the types of patterns that humans find salient in time series data, using natural language descriptions as the only source of supervision during training.
Based on the novel dataset we collect (described in Section \ref{sec:data} 
, we find that the patterns humans identify tend to describe increasing or decreasing trends, volatility, comparisons of start and end values, presence of peaks and dips. They also mention the temporal location of patterns, such as `at the beginning'
of the time series.
Thus, our model should be able to learn patterns such as `increase' or `ends with higher value compared to start', and temporal aspects such as `begin' or `end'.

One way to operationalize this process is through the lens of formal logic: e.g. an increasing trend at the beginning of a time series $x$ 
can be represented trough the logic $z$: $\big[ \exists_i$ s.t. \textsc{increase}($x_i$) AND \textsc{begin}($i$) $\big]$ 
Thereafter, if the program returns \texttt{true} on the input, one can condition on only the logical program $z$ to generate output text that describes this pattern via a decoder, $p(y|z)$. However, this still requires learning or defining modules for patterns and temporal location. Inspired by neural module networks \cite{andreas2016learning,andreas2016neural}, we propose to use functions parameterized by neural networks (Figure \ref{fig:ts_method_overview}) as modules, incorporating inductive bias through architecture design. However, unlike past work, we condition only on an encoding of sampled programs that return \texttt{true} to generate output text. 

\subsection{Model}

Our goal is to generate a text caption $y$
describing a salient pattern in an input time series $x$. Our model’s generative process is depicted in Figure \ref{fig:ts_method_overview} and operates as follows: Conditioned on an input time series $x$, we first sample a program $z$ from a learned prior, $p(z|x)$. 
The latent program $z$ is composed of several operations/modules composed together, and outputs a truth value score. The prior is governed by the truth-values of corresponding programs so that we are likely to sample programs with high truth values.
Next, we sample caption $y$ conditioning \emph{only} on the encoding of sampled program $z$ to generate the final text -- i.e. $y$ is independent of $x$ given $z$. Intuitively, if the latent program encodes sufficient information to describe the pattern it detects, caption needs to only depend on the program itself.

The set of latent `programs' in our model are learned from data. On executing a program $z$ on the input time series data $x$, we obtain an output score $s_z(x)$ (between 0 and 1, both inclusive). Score $s_z(x)$ represents the model's confidence about whether the pattern corresponding to the program holds true for the given input time series. Note that $s_z(x)$ does \emph{not} represent the prior probability of program $z$ -- since multiple programs can be true for a given time series, and $\sum_z s_z(x) \neq 1$.  
% Programs from Modules
We provide our model with a set of building blocks/modules, which combine to form programs. The composition of modules into programs as well as the module parameters are unobserved in data and are learned during model training. The compositionality in the program space enables modules to be shared across programs, leading to more efficient learning. 
The programs we consider will prove quite effective in experiments, but are actually relatively simple, being composed of only three module types. Our framework is extensible, however, and future work might consider larger program spaces.
We refer to our proposed method as \tsmethod{} (\textbf{TRU}th \textbf{C}onditional g\textbf{E}neration).

\subsection{Programs and Modules}
As previously mentioned, each program $z$ in our model is composed of several learnable operations/modules. 
Following prior work on neural modular networks \cite{andreas2016neural}, we consider multiple module types, and incorporate inductive biases in their architecture to learn useful numerical patterns. In the current study, however, we limit to three simple types of patterns: \emph{pattern}, \emph{locate}, and \emph{combine}, leaving extensions to the module space as a future direction.
These modules are composed together into programs that operate on the input time series (Figure \ref{fig:ts_method_overview})

The module types \emph{pattern}  and \emph{locate}, output a vector of the same length as the input vector. Both of them output a temporally localized vector, with each value between 0 and 1 (achieved by applying a sigmoid activation function), representing the degree of confidence that the pattern it represents is present at the corresponding position on the temporal axis. For example, as shown in Figure \ref{fig:vizoutput}, the output of a learned \emph{locate} module is a vector with high values in the middle part, and the output of the \emph{pattern} module is high on those positions where there is a decrease in the value in the input time series.

For the current study, we restrict the space of programs to consist of one \emph{pattern} ($z_P$) module instance and one \emph{locate} ($z_L$) module instance. Outputs from the two modules are combined using a \emph{combine} module, which carries out position-wise multiplication of outputs from $z_P$ and $z_L$, followed by a feed-forward layer and a sigmoid non-linearity. 

\emph{Pattern} modules are aimed at learning patterns such as peaks, dips, increasing trend, and so on. 
We realize \emph{pattern} modules through multi layer 1-D convolutions. We argue that 1D convolutions provide an appropriate architecture to induce aspects such as slopes, and compose them to identify patterns such as peaks. 
The \emph{locate} module types are realized though a mixture model of K fixed Gaussians placed at equal intervals on the temporal axis of given length $T$. The weights of the components represent learnable parameters for such types of modules. 
The \emph{combine} module type learns to transform the position-wise multiplied outputs to a real-valued score, which is then passed through a sigmoid function.

% update: added prior
\subsection{Prior}
As discussed above, the output of each program $z$ is a real-valued score between 0 and 1. We define prior over the set of programs $Z$ as $p(z) \propto e^{\lambda s(z)}$, where $\lambda$ is a hyperparameter.
This formulation makes an implicit assumption that a program $z$ being true for an input time series will make other programs less probable through conservation of probability mass. Such an assumption is necessary, as otherwise directly trying to optimize the likelihood without normalizing across programs will lead to trivial solutions, wherein each program will output a high score for every input. 
Note that an alternative formulation could directly use softmax on an unrestricted real-value output from modules -- such a formulation loses out on the semantics of soft truth output from the programs, and also fared worse in our preliminary experimental evaluations in comparison with the proposed formulation.

\subsection{Decoder}
As mentioned previously, our decoder conditions only on the program $z$ sampled from  the prior $p(z|x)$ to generate final text. 
To achieve this, we need to pass a program representation to the decoder. We consider an auto-regressive neural decoder such as LSTM or Transformer. At every step, the decoder considers embedding of the previous token as well as the input program representation. 

A straightforward approach to obtain program representation is to associate each unique program with a low dimension embedding vector. However, such an approach will not fully exploit the program structures and shared modules. 
Instead, we first associate each module with an embedding. Next, the representation of a program is constructed by appending the embeddings of the corresponding modules (using a fixed pre-determined order of module types). Such a representation achieves sharing of module embeddings across programs. Moreover, it enables obtaining the representation of a new (unseen) program composed using the same set of modules.

%%%%%%%%%%%%%%%%%%%%%%%%%%%%%%%%%%%%%%%%%%%%

\section{Learning and Inference}

The log probability of observing a natural language description $y$ of the time series $x$ under the model can be written as follows:
\begin{equation*}
    \log p(y|x) = \log \sum_{z \in \mathcal{Z}} p_\phi(z|x)p_\theta(y|z)
\end{equation*}
where $\mathcal{Z}$ is the set of all possible programs, and  $\theta$ and $\phi$ are learnable model parameters.
The model is trained to maximize the log likelihood of the the observed descriptions conditioned on the corresponding time series data. Since the programs $z$ are unobserved at training, we must marginalize over all possible values of $z$.
\vspace{2mm}

\noindent \textbf{Inference Network:}
The space of programs we currently employ is relatively small (about 20-60 number of programs), which makes it feasible to marginalize over the program space. However, any future work expanding the space of programs might run into feasibility issues when computing the exact likelihood. In such cases, we can perhaps resort to variational learning to optimize a lower bound to the likelihood by drawing samples from an inference network. 

Additionally, use of inference networks can provide a useful inductive bias by using the observed text descriptions to guide the model learning. For example, words `increase' and `begin' in a caption could inform the inference network about a high chance of the presence of an increase pattern in the initial duration of the time series. 
We observe that training with inference networks results in models which can better capture the patterns in data. 
Note that the inference network is used only for model training. At test time, we sample from the learned prior and decoder without regard to the inference network.

We use amortized variational learning by introducing an inference network $q_\gamma$, and train the model to maximize the following evidence lower-bound (ELBO):
\begin{align*}
    \mathbb{E}_{z \sim q_\gamma(z|y)} [\log p_\theta(y|z)] - \text{KL}(q_\gamma(z|y)||p_\phi(z|x))
\end{align*}

% \noindent \textbf{Architecture:}
We use a BiLSTM encoder to encode the caption $y$, followed by a classification layer to predict the approximate posterior $q_\gamma(z|y)$ over the programs. %module ids for each of the two module types discussed above. 
We also considered fine-tuning of a pre-trained BERT model instead of BiLSTM, but did not observe any improvement in the model performance during the initial experiments.
\vspace{2mm}

\noindent \textbf{Optimization:}
$\theta$, $\phi$ and $\gamma$
%$\psi$
are learned through directly optimizing the ELBO term. 
We compute the exact reconstruction and the KL-terms -- the number of programs in our case is small enough to enable this exact computation (typically we consider 6-10 instances each of \emph{pattern} and \emph{locate} module). 
% Future work: train without need to run all programs

\section{Datasets}
\label{sec:data}

We are interested in modeling numerical patterns and trends in time series data. However, there is a lack of existing data sources with time series data paired with natural language descriptions. 
% 1.
Some prior work on weather forecasting data (such as Sumtime-Mausam \cite{sripada2003sumtime}) are typically small (only 1045 data instances), and are limited in the scope of patterns they encompass.
% 2.
ToTTo dataset \cite{parikh2020totto} contains a small fraction of descriptions based on numerical reasoning and patterns - however, the main challenge is to find the correct value(s) by identifying the relevant row and column in a table.
% 4.
LOGIC-NLG \cite{chen2020logical} consists of 37K tables and corresponding natural language descriptions, some of which require comparisons of cells in a table. 
In contrast, we focus on trends and patterns in time series data. 
Thus, we construct a new dataset where natural language descriptions are collected for naturally occurring stock price time series data (Section \ref{sec:data:stock}). Additionally, we collect natural language descriptions for a synthetically constructed set of time series to evaluate and analyse our models in a more controlled setup (Section \ref{sec:data:synth}).

\subsection{STOCK Dataset}
\label{sec:data:stock}

% Time series data: 
We collect naturally occurring time series data in the form of stock prices. We utilize the Google Finance API to collect stock prices of 7 randomly chosen technology companies over a period of 20 years. We collect weekly (beginning of week) as well as and daily stock price values. 
We sub-select a total of 1900 instances, each of consists of sequence of T(=12) values. 
Each instance is sampled from the stock data as follows: (1) we pick one of the companies uniformly at random (2) we randomly pick weekly or daily series with equal probability, (3) we pick a sequence of values of given length T, ensuring no overlap with any previously selected time series. (4) Additionally, since different company stocks can be in very different range of values, we normalize such that all the values are between 0 and 100: $v' = 100*(v-min)/(max-min) $ . However, normalizing this way directly would create undesirable biases in the dataset since each time series would necessarily cover entire range 0-100. Instead, to compute \emph{max} and \emph{min}, we additionally consider 10 values (chosen based on manual inspection) just before and just after the currently selected range. \vspace{2mm}

% Stock data: 
\noindent \textbf{Annotation collection:}
We collect 3 natural language annotations for each of the 1900 data points, leading to a total of 5700 paired time-series with natural language descriptions. We split the 1900 unique time series and associated captions into train, dev, and test splits with ratio 8:1:1.

\noindent \textbf{Annotator description:}
We use Amazon Mechanical Turk as a crowd-sourcing platform.
We limit to annotators from Anglophone countries, with HIT (Human Intelligence Task) acceptance rates of more than $90\%$, and minimum number of accepted HITs as 100. Annotators were paid 25 cents for each annotation (which comes to average hourly rate of over USD 23).

\noindent \textbf{Quality Control:}
Based on initial pilot studies, we found it useful to show annotators plots instead of tables of values, as we are interested in high level patterns rather than specific values. We do not label the plot lines with actual stock names to remove any potential biases one may have about specific company stocks. Finally, we restrict annotations to a maximum of 9 words, so that one annotation reflects only one pattern. % (Annotators was also explicitly told to limit to only one `pattern' per annotation). 
Each HIT is labelled by 3 different annotators. We manually inspected at least one annotation from each unique annotator, and ruled out (but still paid) annotations for about 7\% annotators for being poor quality.

\noindent \textbf{Encouraging Lexical Diversity:}
We encouraged annotators (through instructions) to not limit themselves to words shown in examples. Additionally, we limit each annotator to a maximum of 10 HITs to increase diversity in annotations.

\noindent \textbf{Dataset Statistics:}
There are a total of 861 unique words across the 5700 captions. Most annotation sentences follow a simple syntactic structure. Additionally, we picked a random subset of 100 data points, and manually classified most of them into following major buckets: trend (increase/decrease trends: 48\%) superlative(max/min values; peaks and troughs: 20\%); comparisons(comparison of start and end values: 10\%); volatility (flat/smooth; irregular: 12\%).

\subsection{Synthetic Time Series (SYNTH)}
\label{sec:data:synth}
To develop and test models in a more controlled setup, we synthetically construct time series data. 
Our synthetic time series data is constructed such that each time series has exactly one of the following 6 patterns: increases-in-beginning, increases-in-middle, increases-in-end, decreases-in-beginning, decreases-in-middle, decreases-in-end. 
The resulting dataset consists of a total of paired 720 time series - natural language annotations.

% \subsubsection{Construction Details:}
Each synthetic time series is generated as follows: 
First, the trend is chosen: increase or decrease. A trend is realized through a straight line of length $L<=T/3$, with randomly chosen intercept and slope within a range based on the trend selected. 
Next, we randomly select one of the 3 temporal locations : begin, middle, end -- and based on the choice, the pattern is placed in first 40 percentile, 30-70 percentile, or 60-100 percentile respectively, of the entire length T. The region outside the trend is flat. 
Finally, small noise is added to each point. The setup is such that the resulting values are always in (0,100) range. Examples and more specific details can be found in Appendix. % \ref{appendix:sec:synth}. 

% \section{Experiment Setup}
\section{Experiments with Synthetic Data}

\begin{table}[t]
    \centering
    \footnotesize
    \begin{tabular}{@{}l@{\hskip 0.05in}l@{\hskip 0.05in}l@{\hskip 0.05in}l@{\hskip 0.05in}l@{\hskip 0.05in}l@{\hskip 0.05in}lll@{}}
        \toprule
        \bf Method & \bf COR & \bf PPL & \bf Bleu-3/4 & \bf Cider & \bf Rouge  & \bf BERT  \\ \midrule
        \tsmethod{} & $\bf 92\%$ & $13.9$ & $0.61/0.46$ & $1.40$ & $0.74$ & $0.77$   \\  
        \seqff{} & $39\%$ & $16.7$ & $0.45/0.28$ & $0.81$ & $0.61$ & $0.65$ \\  
        \seqlstm{} & $45\%$ & $11.2$ & $0.43/0.28$ & $0.87$ & $0.62$ & $0.63$  \\  % 
        \seqconv{} & $53\%$ & $11.0$  &  $0.47/0.32$ & $1.00$ & $0.66$ & $0.67$  \\  
        % \seqconvmulti{} &   \\  
        \seqfft{} & $39\%$ & $22.7$ & $0.38/0.22$ & $0.67$ & $0.58$ & $0.54$  \\  %
        \retrieval{} & $71\%$ & NA & $0.28/0.14$ & $0.60$ & $0.40$ & $0.48$  \\ 
        \bottomrule
        \end{tabular}
        \caption{\footnotesize
        Results on test split of SYNTH dataset: Human evaluation for correctness (COR) and various automated metrics. \tsmethod{} performs much better than baselines as per correctness evaluation. 
        }
        \label{tab:synth6_results}
\end{table}
% \vspace{-0.4\abovedisplayskip}

\subsection{Methods}
For SYNTH data, we consider several baselines listed below (More detailed descriptions are provided in the Appendix). Note that all non-retrieval baselines use the same LSTM decoder architecture as our model. 
(1) \textbf{\retrieval{}:} The ground-truth caption of the closest matching training data instance is used as the prediction. The closest matching instance is identified via L2 distance between input time series. %, and uses its accompanying annotation as prediction. 
(2) \textbf{\seqff{}}: Encodes the input time series sequence using a multi-layer feed-forward encoder. %, followed by an auto-regressive LSTM decoder.
(3) \textbf{\seqlstm{}}: Encodes the input time series sequence using a LSTM recurrent neural network. % (Additional details in appendix). 
(4) \textbf{\seqconv{}}: Encodes time series using a multi layer convolutional neural network. 
(5) \textbf{\seqfft{}}: Encodes time series using Fourier transform features of the input.

\subsection{Results}

For \tsmethod{}, we pick the highest scoring program, according to the prior, for description generation. 
We generate captions (using greedy decoding) from each of the methods for the test split.
\\
\textbf{Automated metrics} measure overlap between model generated caption and the reference ground truth captions. We report Perplexity (\textbf{PPL}), BLEU-3/4 \shortcite{papineni2002bleu}, METEOR \cite{banerjee2005meteor}, ROUGE-L (\textbf{Rouge}) \cite{lin2004rouge}, and BertScore-Precision (\textbf{BERT}) \cite{DBLP:conf/iclr/ZhangKWWA20}. The proposed \tsmethod{} method gets favorable scores as per various automated metrics on the test split of SYNTH (Table \ref{tab:synth6_results}).

\noindent \textbf{Human Evaluations for Correctness:} 
Automated metrics may not correlate well with actual quality of the generated output in text generation tasks \cite{celikyilmaz2020evaluation}.
As such, we report human evaluation results as well. We recruit human annotators who are requested to provide a binary label on factual correctness \textbf{(COR)} of the captions for the test split. Each caption is annotated by three annotators, and the majority label is used. The proposed method is able to achieve a high correctness score of $92\%$, which is much better than the baselines. This demonstrates the usefulness of the proposed truth-conditional model in generating highly faithful captions. 
Output samples are provided in the Appendix.

\begin{SCtable}[]
    % \floatbox[\capbeside]{table}
    \centering
    \footnotesize
    \begin{tabular}{@{}ll@{}}
        \toprule
        \bf Method & \bf COR  \\ \midrule
        \tsmethod{} &  $\bf 97\%$    \\  
        \seqff{} &  $38\%$ \\  
        \seqlstm{} &  $50\%$ \\  
        \seqconv{} & $59\%$ \\
        \seqfft{} &  $39\%$ \\  
        \retrieval{} &  $72\%$ \\  
        \bottomrule
    \end{tabular}
    \label{tab:synthetic_clf_transfer}
    \caption{\footnotesize Models trained on SYNTH data (where each time series has T=12 values) are tested on another synthetic data with T=24 without any fine-tuning.
    }
% \end{wraptable}
\end{SCtable}

\subsection{Analysis}
\noindent \textbf{Generalization to different time series duration:} SYNTH data consists of time series instances with T=12 sequence of values. We experiment the extent to which models trained on SYNTH can accurately detect patterns in time series data of different lengths without any fine-tuning. For this, we evaluate results on a separate synthetic data consisting of 100 time series with T'=24 values per time series (dataset created in the same manner as SYNTH and consists of the same set of 6 classes as in SYNTH).  
% \begin{wraptable}{r}{5.5cm}

We observe that \tsmethod{} retains high correctness of the output captions (Table \ref{tab:synthetic_clf_transfer}), whereas some of the high performing baseline show significant reduction in correctness.
Note that some of the employed methods like \retrieval{} and \seqff{} cannot work directly on inputs of length different than present in the training data. For such models, we first adjust length of series. For example, for length 24 input, we consider alternate values only, thereby reducing the series to length 12 (same as in the training data).  
\vspace{2mm}
%  \\

\begin{table}
\footnotesize
\begin{tabular}{l@{\hskip 0.1in}l}
\toprule
\bf Module  & \bf Most freq. words associated \\
\bf id & \bf with learned modules   \\ \midrule
pattern-1 &  increases, rises   \\
pattern-2 &  decreases, decline, dips \\  
locate-1 &  end, late  \\  
locate-2 &  beginning , start, initial \\
locate-3 &  middle, halfway  \\
\bottomrule
\end{tabular}
  \caption{\footnotesize Some of the most frequent words associated with some of the learned module instances for SYNTH data.
  \label{tab:module_analysis}
}
\end{table}

\noindent \textbf{Analyzing Learned Modules:}
We analyze the characteristics of the learned modules by identifying the top words (excluding stop words) associated with each learned module. To do so, for a given series, we find program with highest score, and associate the annotations for that series to corresponding modules in that program. Finally, we collect the most frequent words in annotations associated with each module. We show a summary in the Table \ref{tab:module_analysis}. The two trend modules seem to be getting activated for increase and decrease patterns respectively.
\vspace{2mm}

\noindent \textbf{Compositionality of Learned Modules}
We analyze if the proposed model uses its compositional parameterization effectively.
To do so,
we conduct a simple analysis as follows: % experiment: %two experiments.
% (1) 
We train \tsmethod{} on a subset of synthetic data consisting of only the following 4 patterns: increase-beginning, decreases-end, increase-middle, decreases-middle. 
We examine this trained model's behavior on test data points consisting of the two unseen patterns: increase-end and decrease-beginning. More specifically, we analyze the argmax program prediction as per the conditional prior. Based on manual inspection of modules (similar to what we discussed for analysis in Table \ref{tab:module_analysis}), we know before hand the program which should be selected for these patterns. Model's prediction is considered to be correct if, for example, for an input with `decrease-beginning' pattern, model assigns highest score to the program composed using modules corresponding to `decrease' and `beginning'.
We observe that the highest scoring program is the correct/expected program for 92\% of the cases in the test split.

\section{Experiments with STOCK Dataset}

\subsection{Posterior Regularization:}
In the initial experiments with STOCK dataset, we observe that our model suffers from model collapse, and degenerates into learning a single program only. 
This is perhaps because randomly initialized modules do not 
have much guidance to begin with. To mitigate such mode collapse issues, prior work has used mutual posterior divergence (MPD) regularization  \cite{ma2019mae} 
$-E_{y_i,y_j} KL(q(z|y_i)||q(z|y_j)) $,
where $y_i$ and $y_j$ captions for two randomly chosen data points. 

However, we note that MPD term enforces the divergence in an indiscriminate manner -- divergence is encouraged even if captions are paraphrases of each other. An alternate way to encourage divergence in the inference network prediction is to encourage divergence only when two captions $y_i$ and $y_j$ represent different programs or patterns. However, such information is not available in the training data. 
Instead, we use an approximation as follows: 
We identify the $M$ most frequently occurring words excluding stop-words (list available in Appendix) in the captions and are manually labelled to to represent pattern or locate or neither. Each of the words labelled to be of type pattern or locate is assigned a unique \emph{pattern} or \emph{locate} module id respectively. 
The corresponding captions thus get tagged with some heuristic (but potentially noisy) labels for module ids.
Only those captions are tagged which have exactly one `locate' word and one `pattern' word.
This leads to about 31\% of the captions being assigned such heuristic labels, while the remaining data stays unlabelled. 

The above procedure does involve a small human-in-the-loop component.  However, we note that it is a pretty light-weight involvement. For example, the system presents M(=10) most frequent pairs of words (excluding stopwords) in captions, and a person spends a couple of minutes labeling their type (locate or pattern).

\subsection{Results}
We now report results with STOCK dataset.
As mentioned above,
we utilize heuristic labels as an auxiliary loss when training the proposed method. Thus, for a fair comparison,
the baselines \textbf{\seqlstmmulti{}}, \textbf{\seqconvmulti{}} and \textbf{\seqffmulti{}} also use the same set of heuristic labels via a classification loss on the encoded representation in a multi-task learning setup.
 
The proposed method \tsmethod{} produces high precision captions as judged by human annotators (Table \ref{tab:stock_results}). 
We additionally report automated text overlap scores against reference captions, though the automated metrics seem only mildly correlated with human judgement ratings. 
Interestingly, some of the baselines show large differences in performance in STOCK vs SYNTH datasets. For example, \retrieval{} performs well on SYNTH but rather poorly on STOCK dataset, perhaps because of variety in time series instances in SYNTH being small, while the same being large in STOCK.
\vspace{2mm}

% \subsection{Diversity and Coverage}
\noindent \textbf{Diversity and Coverage:}
Ideally, we want models which can  identify all the interesting patterns present in an input time series. Correctness results discussed earlier are indicative of faithful generation but do not necessarily capture coverage of patterns.
We compute coverage of various models via the following procedure. First, we collect L(=12) samples per data point from the model. Next, we recruit human annotators to rate whether a human written reference annotations for that data point is covered by the set of L generated captions or not.
For \tsmethod{}, we perform sampling at the program selection stage, while baselines admit sampling only at the token generation stage.

\begin{table}[]
    \centering
    \footnotesize
\begin{tabular}{@{}l@{\hskip 0.05in}l@{\hskip 0.09in}l@{\hskip 0.09in}l@{\hskip 0.05in}l@{\hskip 0.05in}l@{\hskip 0.05in}ll@{}}
\toprule
\textbf{Method} & \textbf{COR} & \textbf{Bleu-3/4} & \textbf{Cider} & \textbf{Rouge} & \textbf{BERT}  \\
\midrule
\tsmethod{}(Ours) & \bf 88.4\% & 0.35 / 0.19 & 0.36 & 0.50 & 0.57  \\
\seqffmulti{} & $64.2\%$ & 0.32 / 0.19 & 0.43 & 0.47 & 0.56  \\
\seqlstmmulti{} & $65.5\%$ & 0.35 / 0.21 & 0.41 & 0.50 & 0.61 \\
\seqconvmulti{} & $65.9\%$  & 0.33 / 0.18 & 0.41 & 0.49 & 0.59 \\
\seqfft{} & 61.8\% &  0.34 / 0.19 & 0.39 & 0.49 & 0.58 \\
\retrieval{} & 47.2\%  & 0.12 / 0.06 & 0.14 & 0.28 & 0.35  \\
\bottomrule
\end{tabular}
    \caption{\footnotesize Results with STOCK data: Proposed method \tsmethod{} scores the best on  correctness evaluation. The best performing baseline scores $20\%$ less on correctness evaluation. Greedy decoding was used for all the methods.}
    \label{tab:stock_results}
\end{table}

\begin{figure}[t]
\begin{center}
    \includegraphics[width=0.65\textwidth]{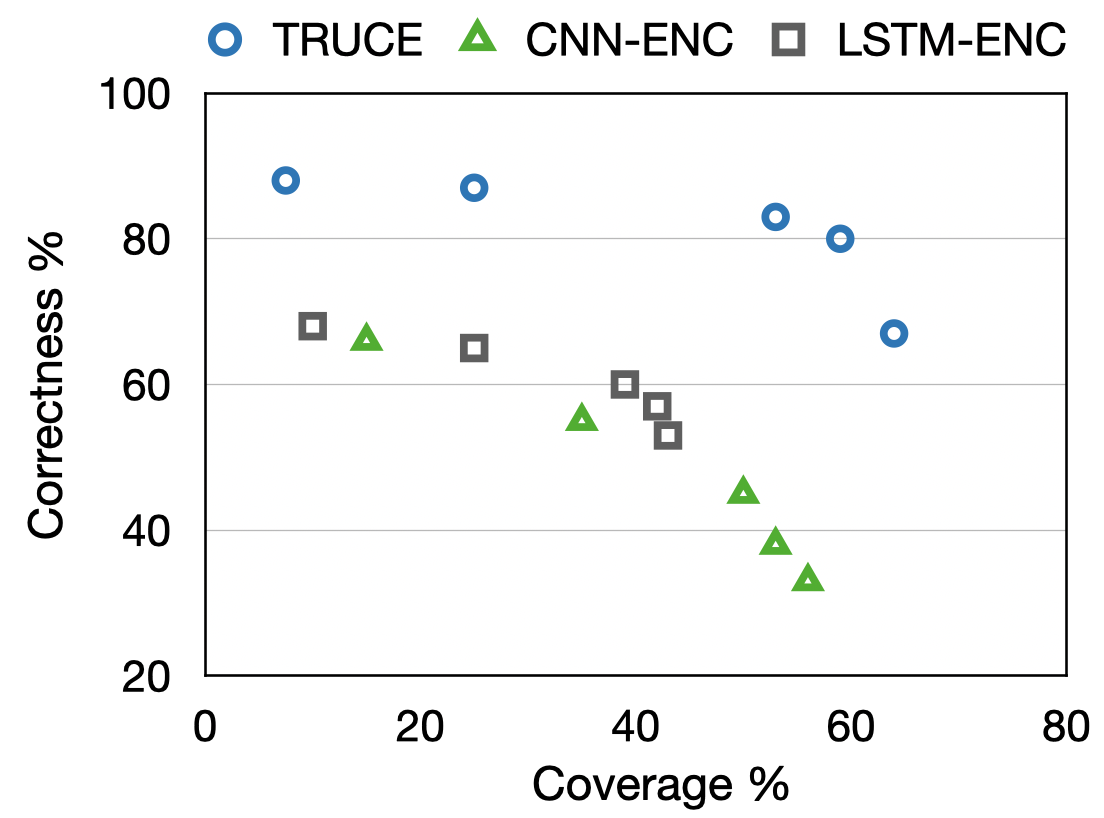}
    \vspace{-0.5\abovedisplayskip}
    \caption{\footnotesize Coverage and Correctness of model outputs at different sampling settings. In general, settings with higher coverage of human written captions have lower precision of generated captions. \tsmethod{} achieves much higher correctness scores compared to baselines for similar coverage values.}
    \label{fig:coverage}
    \end{center}
    \vspace{-3mm}
\end{figure}
Note that this makes the coverage score depend on the settings used in the sampling process (e.g. top-p value in nucleus sampling), which will also affect the correctness of the generated captions. In Figure \ref{fig:coverage}, we demonstrate coverage and correctness values of \tsmethod{} and two of the baseline models under different sampling conditions. In general, restricting samples to a low value of top-p leads to lower coverage but higher correctness. 
Overall, \tsmethod{} behaves in a more favorable manner. For example, comparing \tsmethod{} against \seqconv{}, for roughly same level of coverage (e.g. ~50\%), correctness is much higher for \tsmethod{} (~83\% against ~45\% for \seqconv{}). 
However, there still seems to be a gap in the coverage of patterns, and can perhaps be addressed by incorporating more module types. 
\vspace{2mm}

\subsection{Analysis}

\noindent \textbf{Direct conditioning on the input:}
Our decoder conditions only an encoding of a sampled program. We hypothesize that such an approach creates a bottleneck discouraging the decoder from learning spurious correlations between the input time series and the output text. 
To inspect the usefulness of the proposed abstraction, we consider an alternative model wherein the decoder conditions on the input time series as well -- by providing output of a convolutional encoder (same as in  \seqconv{}) to the decoder. More specifically, the program representation and the encoder representation are concatenated before being fed to the decoder. Lets refer to such a model with decoder having direct access to the input as \tsmethod\textsc{-D}. For STOCK data, \tsmethod\textsc{-D} gets correctness of $69\%$ compared to $88\%$ for \tsmethod{}.  \\

\noindent \textbf{Analysis of Inference Network:}
We analyze the predictions of the inference network at the end of model training.  Particularly, we associate the set of ground truth annotations in validation split to module-ids present in the argmax program prediction from the inference network. Next, we identify the most frequently occurring tokens present for each module-id/module-instance. We observe that the inference network seems to be associating semantically similar words to the same module instance (\autoref{tab:inference}). 
\begin{table}[]
    \centering
    \footnotesize
    \begin{tabular}{l@{\hskip 0.1in}l}
    \toprule
    \bf Module id  & \bf Most frequently associated words\\
    \midrule
    pattern-1 &  increases, rises, gains \\
    pattern-3 &  stays, remains, flat  \\
    pattern-4 &  bottoms, out, decline, dips 
    \\  
    loc-1 & start, beginning, initially  \\  
    \bottomrule
    \end{tabular}
    \caption{\footnotesize Inference Network Analysis: Analyzing words frequently present in captions when the argmax program prediction from inference network comprises of a give module-id.}
    \label{tab:inference}
\end{table}

\section{Related Work}

\noindent \textbf{Time-Series Numerical Data and Natural Language}
\newcite{andreas2014grounding} worked on grounding news headlines to stock time series data by aligning sub-trees in sentence parses to segments of time series. 
\newcite{DBLP:conf/acl/MurakamiWMGYTM17} generate stock data commentary using encoders such as convolutional and recurrent neural networks, similar to the baselines used in our experiments. 
\newcite{sowdaboina2014learning} focus on the task of describing wind speed and direction. 
Time series data in the form of charts has been utilized in some prior work in figure question answering \cite{DBLP:conf/iclr/KahouMAKTB18,DBLP:journals/corr/abs-1906-02850}.

Past work has explored ways to handle numerical data in a variety of input data domains using neural networks.
\newcite{trask2018neural} propose neural logic unit for tasks such as counting objects in images. 
Prior work has investigated handling of numeracy in question answering datasets \cite{dua2019drop,andor2019giving,DBLP:conf/iclr/GuptaLR0020}, typically using a predefined set of executable operations or using specific distributions for number prediction \cite{DBLP:conf/emnlp/Berg-Kirkpatrick20,DBLP:conf/naacl/ThawaniPIS21}.

\noindent \textbf{Neuro-Symbolic Methods:}
\citet{andreas2016neural} proposed to use neural modular networks for visual question answering. Since then, similar approaches have been used for several other tasks such as referring expression comprehension \cite{DBLP:conf/aaai/CirikBM18}, image captioning \cite{DBLP:conf/iccv/YangZC19}, and text question answering {\cite{andreas2016learning,DBLP:conf/naacl/KhotKRCS21}}. %
Compared to such past efforts, we induce the latent numerical and temporal detection operations, pick a high-scoring program, and condition only on a program encoding to generate the output description. 
In this respect, our work is also related to prior work on neural discrete representation learning \cite{DBLP:conf/nips/OordVK17,DBLP:conf/acl/EskenaziLZ18}, though none of these past works explore utilizing such techniques for data to text problems. 
Our proposed model abstracts the numerical pattern detection from text generation. Related ideas have been explored in the past in other domains and tasks \cite{DBLP:conf/emnlp/GehrmannDR18,DBLP:conf/emnlp/JhamtaniB18,DBLP:conf/icml/AmizadehPPHK20}.

\noindent \textbf{Data to Text:}
Tabular or structured data to text generation has been explored in prior work \cite{lebret2016neural,DBLP:conf/sigdial/NovikovaDR17,wiseman2017challenges,jhamtani2018chess,DBLP:journals/corr/abs-2102-01672}. 
The Rotowire dataset \cite{wiseman2017challenges} is comprised of sports summaries for tabular game data which may require modeling of numerical operations and trends. % to identify gaming concepts such as `defeated'. 
However, much of the past work has relied on neural models with attention mechanisms, without explicit and interpretable notions of numerical operations.
Fidelity to the input in the context of neural text generation has received a lot of attention lately \cite{DBLP:conf/aaai/CaoWLL18}. 
Prior work has approached the aspect of fidelity to input through changes in model training and/or decoding methods \cite{DBLP:journals/corr/abs-1910-08684,DBLP:conf/acl/KangH20,DBLP:conf/acl/MajumderBMJ20,DBLP:conf/naacl/GoyalD21,DBLP:conf/aaai/0001ZCS21}. 
We explore a different approach that increases fidelity through conditional independence structure and model parameterization.

\section{Conclusion}
We present a truth-conditional neural model for time series captioning. Our model composes learned operations/modules to identify patterns which hold true for a given input. Outputs from the proposed model demonstrate higher precision and diversity compared to various baselines. Further, the proposed model (and some of the baselines) successfully generalize, to some extent, to multiple input sizes. We release two new datasets (in English) for the task of time series captioning. Future work might expand to a broader set of module types to cover more numerical patterns. 

\section*{Acknowledgements}
We thank anonymous EMNLP reviewers for insightful comments and feedback. We thank Nikita Duseja for useful discussions.

%\newpage
\section*{Ethics Statement}
We collect natural language annotations from a crowd-sourcing platform. We do not collect or store any person identifiable information. We did not observe any toxic or hateful language in our dataset -- though researchers working on the dataset in future are advised due caution since the annotations are crowd-sourced, and might reflect certain biases.
Our work primarily performs experiments on text generation in English language. 
Our method generates high precision text output -- much higher than all the baselines considered. However, it is still not perfect, and must be used cautiously in any real world deployment.

\bibliography{main.bib} % anthology.bib,
\bibliographystyle{acl_natbib}

\clearpage
\appendix

\section{Additional Details on Data Sets}

A downloadable json file for each of the two datasets is provided in the github repository \footnote{\url{https://github.com/harsh19/TRUCE}}.

\subsection{Synthetic Data}
\label{appendix:sec:synth}

% \noindent \textbf{Data Collection:}
Our synthetic time series data is constructed such that each time series has exactly one of the following 6 patterns: increases-in-beginning, increases-in-middle, increases-in-end, decreases-in-beginning, decreases-in-middle, decreases-in-end. 
The position in which the pattern is placed is based on the temporal choice (begin/middle/end). i.e. L must lie withing first one-third of the time-series (0,T/3) in case of `begin' pattern, should lie in middle one-third for `middle', and last one third for `end' respectively. We consider equation a*x+b of a line, where `a' represents the slope and `b' represents the y-axis intercept.  We pick a random slope value between 0 and 2, and a random intercept value between 1 and 20. Finally, we pick $|L|$ random integral values for x such that ax+b point lies between 0 and 1. The points in the time series outside the pattern are fixed to be same as the nearest point in the patter. Finally, small noise is added to each point using U(-2,2).  

Some random data samples are shown in Fig. \ref{fig:syntheg1}. The text corresponding to `HUMAN' marker represents one of the collected annotations for the corresponding time series data.

\subsection{STOCK data}

% \noindent \textbf{Data Samples:}
Figures \ref{fig:stockeg1} show data samples for STOCK dataset. The text corresponding to `HUMAN' marker represents one of the collected annotations for the corresponding time series data. 
The total number of unique words (considering train and validation splits) are 861, out of which only 560 words occur more than once in the dataset.

\section{Additional Results}

\subsection{SYNTH: Generated Samples}
\label{appendix:sec:synthgensamples}
\begin{figure*}[t]
    \centering
    \includegraphics[width=0.85\textwidth]{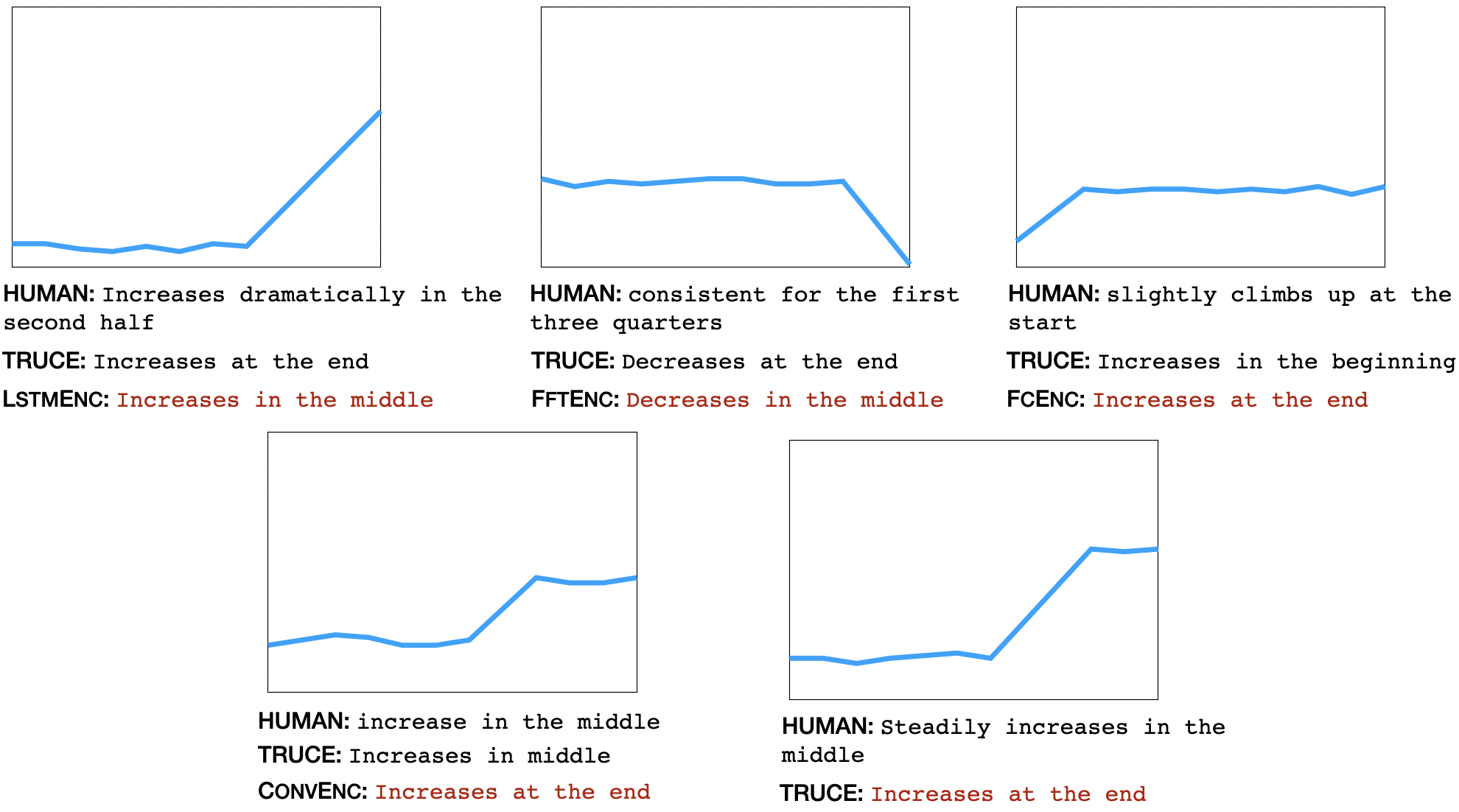}
    \caption{SYNTH: Data and Generated Samples. The captions marked in red were judged as incorrect by human annotators. \tsmethod{} achieves very high precision of 95\% on outputs for the test split of SYNTH dataset. }
    \label{fig:syntheg1}
\end{figure*}

Additional examples are provided in Figure \ref{fig:syntheg1}.

\subsection{STOCK: Generated Samples}
\label{appendix:sec:stockgensamples}
\begin{figure*}[t]
    \centering
    \includegraphics[width=0.85\textwidth]{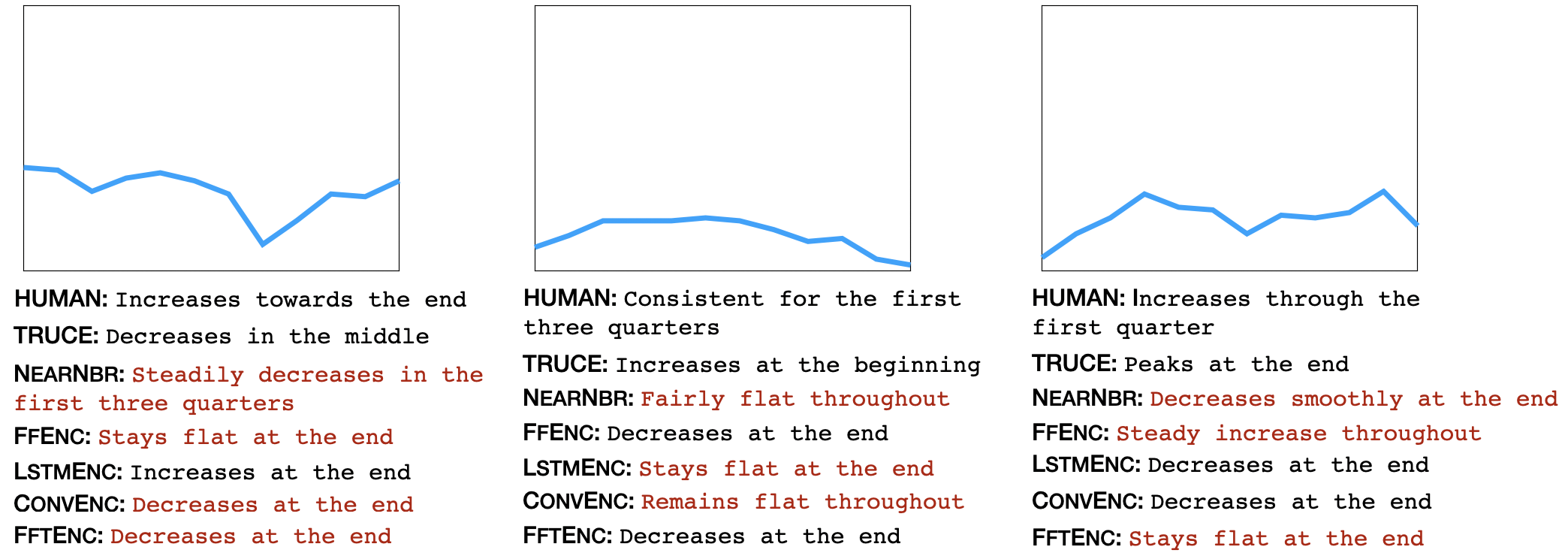}
    \caption{STOCK: Data and Generated Samples. The captions marked in red were judged as incorrect by human annotators. (Best viewed in color)}
    \label{fig:stockeg1}
\end{figure*}
Figure \ref{fig:stockeg1} shows some generated samples on STOCK dataset.

\subsection{Validation Split Results}
Tables \ref{tab:synth6_results_val} and \ref{tab:stock_results_val} show automated metrics on the validation split.

\begin{table}[]
    \centering
    \footnotesize
    \begin{tabular}{@{}l@{\hskip 0.05in}l@{\hskip 0.05in}l@{\hskip 0.05in}l@{\hskip 0.05in}l@{\hskip 0.05in}l@{\hskip 0.05in}lll@{}}
        \toprule
        \bf Method & \bf PPL & \bf Bleu-3/4 & \bf Cider & \bf Rouge  & \bf BERT  \\ \midrule
        \tsmethod{}  & $9.02$ & $0.61/0.50$ & $1.92$ & $0.74$ & $0.76$   \\  
        \seqff{}  & $9.66$ & $0.41/0.34$ & $1.17$ & $0.63$ & $0.57$ \\  
        \seqlstm{}  & $7.5$ & $0.43/0.35$ & $1.39$ & $0.63$ & $0.63$  \\  
        \seqconv{}  & $7.6$  &  $0.63/0.53$ & $1.99$ & $0.73$ & $0.71$  \\  
        \seqfft{}  & $15.7$ & $0.39/0.29$ & $1.26$ & $0.61$ & $0.62$  \\  %
        \retrieval{} & NA & $0.32/0.19$ & $0.68$ & $0.50$ & $0.48$  \\ 
        \bottomrule
        \end{tabular}
        \caption{\footnotesize
        Results on validation split for SYNTH dataset. 
        }
        \label{tab:synth6_results_val}
\end{table}

\begin{table}[]
    \centering
    \footnotesize
\begin{tabular}{@{}l@{\hskip 0.05in}l@{\hskip 0.09in}l@{\hskip 0.09in}l@{\hskip 0.05in}l@{\hskip 0.05in}l@{\hskip 0.05in}ll@{}}
\toprule
\textbf{Method} & \textbf{Bleu-3/4} & \textbf{Cider} & \textbf{Rouge} & \textbf{BERT}  \\
\midrule
\tsmethod{}(Ours) &  0.36 / 0.22 & 0.40 & 0.50 & 0.58  \\
\seqffmulti{}  & 0.32 / 0.20 & 0.38 & 0.47 & 0.56  \\
\seqlstmmulti{} & 0.34 / 0.18 & 0.33 & 0.51 & 0.61 \\
\seqconvmulti{}  & 0.34 / 0.17 & 0.35 & 0.50 & 0.60 \\
\seqfft{}  &  0.32 / 0.18 & 0.36 & 0.48 & 0.56 \\
\retrieval{}  & 0.11 / 0.05 & 0.11 & 0.27 & 0.37  \\
% \midrule
% \unconlm{} & \\  
\bottomrule
\end{tabular}
    \caption{\footnotesize Results on validation split of STOCK data.
    }
    \label{tab:stock_results_val}
\end{table}

\subsection{Analyzing Learned Modules}
\begin{figure*}[]
    \includegraphics[width=0.75\textwidth]{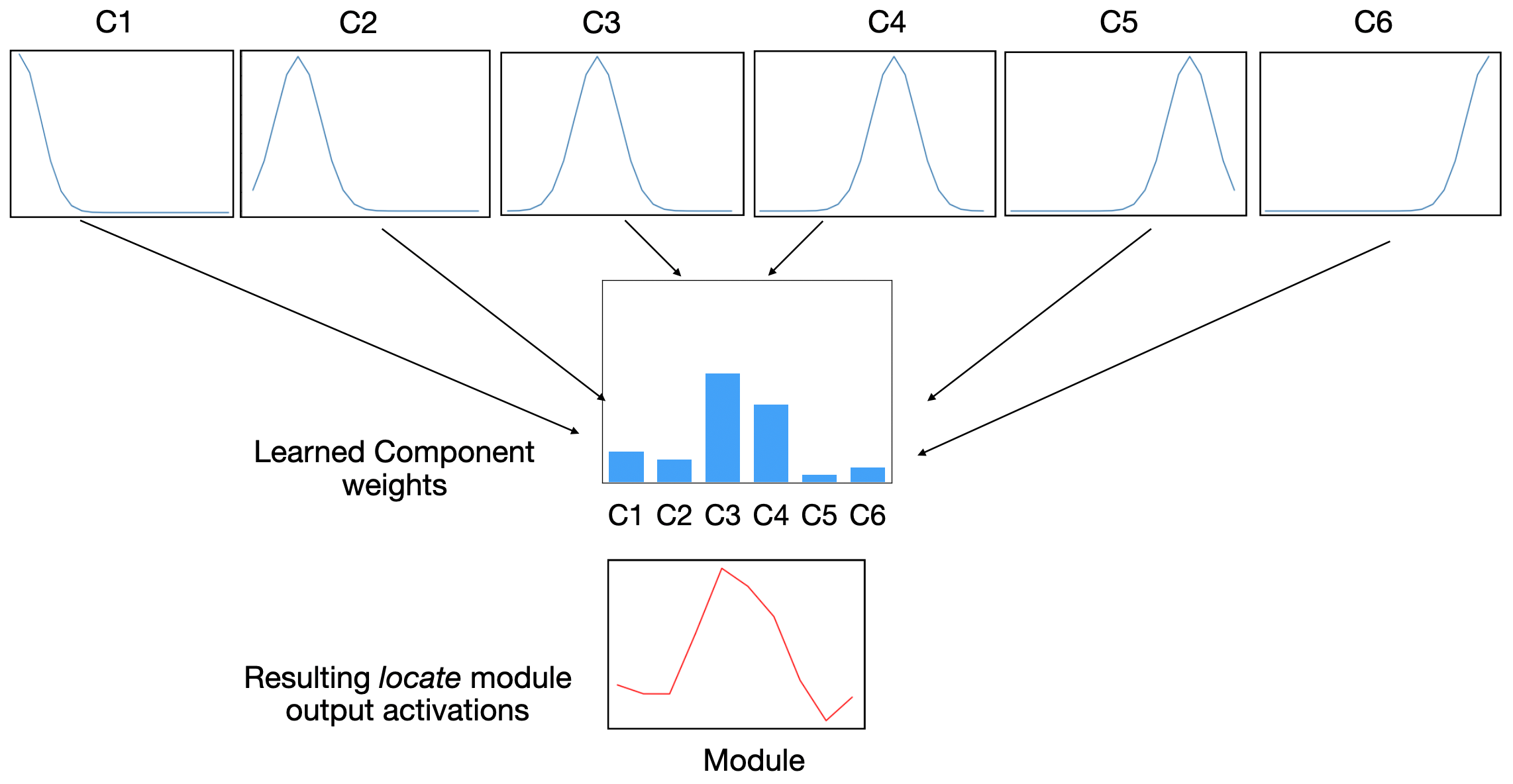}
    \caption{Visualizing a learned 'locate' module. Our locate modules are weighted mixtures of equally spaced Gaussians. The module's weight on each of these components is shown, along with the resulting distribution -- the module being visualized seems to have learned to focus on middle part of the time series.}
    \label{fig:loc_module_viz}
\end{figure*}

Figure \ref{fig:loc_module_viz} shows visualization of a learned \emph{locate} module when model is trained on SYNTH data.

\subsection{Additional Ablation Studies}
We consider following ablations for the \tsmethod{}:
(1) \tsmethod\textsc{-NoInf}: Train \tsmethod{} without the use of inference network
(2) \tsmethod\textsc{-NoHeur}: Train \tsmethod{} without the use of heuristic labels

\section{Additional Training Details}

We code our models in Pytorch library. 

\subsection{Heuristic Labels}
List of the keywords selected for use in constructing heuristic labels: \\
--- `locate':[`beginning',`middle',`end',`throughout'], \\ 
--- `pattern':[`increase',`decrease',`peak',`flat',`dip']

\subsection{Optimizer}
We use Adam optimizer with initial learning rate of $1e-4$.

\subsection{Infrastructure}
We use GeForce RTX 2080 GPUs for training models.

\subsection{Additional method details} 
While the automated metrics are only moderately correlated with quality, we found it reasonable to select best model configurations based on the Bleu-4 scores on validation split. 
The model configurations, when using STOCK dataset, are as follows:
\begin{itemize}
    \item LSTM Decoder: Token embedding size and  hidden size are varied from the set \{32,64,128,256\}. 
    \item Weight for the classification loss term (in case of multitask objective in baselines): Following three weights of classification loss (i.e. the weight of the classification term which is present in addition to the conditional language modeling objective) are tried: 0.3,1.0,3.0. 
    \item \tsmethod{}: Program embedding encoding size. Number of module instantiations are varied in following ranges:
    \begin{itemize}
        \item LOCATE: 4-7 instantiations of each of locate 
        \item PATTERN: 6-10 instantiations of each of trend
        \item COMBINE: 1 instantiation
    \end{itemize}
    - Module embedding is varied in the set \{9,18,36,72\}. Final module embedding size is 18. \\
    - Number of trainable parameters: 466K (excluding inference network parameters since inference network is used only at training and not at prediction time)
    \item \seqfft{}: 
    - Number of trainable parameters: 462K
    - Construct features based on numpy:fft:rfft functions, using real as well as imaginary components from the transformation.
    \item \seqconv{}: 
    Number of trainable parameters: 463K
    \item \seqlstm{}: 
    - Representation: A single LSTM step involves feeding an embedding of the input and using the previous step's hidden state. To construct an input embedding of size $h$ for a given number $x_t$, we simply repeat the number $x_t$ for $h$ times. \\
    - Number of trainable parameters: 464K 
     \item \retrieval{}: 
     We experiment with L2 distance and L1 distance, and observed former to perform better in terms of automated as well as human evaluations. 
\end{itemize}

\end{document}